\documentclass{article}

\usepackage[nonatbib,final]{nips_2017}

\usepackage[utf8]{inputenc} 
\usepackage[T1]{fontenc}    
\usepackage{hyperref}       
\usepackage{url}            
\usepackage{booktabs}       
\usepackage{amsfonts}       
\usepackage{nicefrac}       
\usepackage{microtype}      
\usepackage[sort,numbers]{natbib}

\usepackage[export]{adjustbox}
\usepackage{subfig}
\usepackage{graphics} 
\usepackage{graphicx}
\graphicspath{{./figures/}}

\usepackage{color}

\title{Automatically Extracting Action Graphs from Materials Science Synthesis Procedures}

\author{
  Sheshera Mysore$^1$ \qquad Edward Kim$^2$ \qquad Emma Strubell$^1$ \qquad Ao Liu$^1$  \\
  {\bf Haw-Shiuan Chang$^1$ \qquad Srikrishna Kompella$^1$ \qquad Kevin Huang$^2$} \\
  {\bf Andrew McCallum$^1$ \qquad Elsa Olivetti$^2$}\\
  $^1$College of Information and Computer Sciences\\
  University of Massachusetts Amherst\\
  \texttt{\{smysore, strubell, aoliu, hschang, skompella, mccallum\}@cs.umass.edu}
  \vspace{0.25cm} \\
  $^2$Department of Materials Science and Engineering\\
  Massachusetts Institute of Technology\\
  \texttt{\{edwardk, kjhuang, elsao\}@mit.edu}
  \\
}

\begin{document}

\maketitle

\begin{abstract}
  Computational synthesis planning approaches have achieved recent success in organic chemistry, where tabulated synthesis procedures are readily available for supervised learning. The syntheses of inorganic materials, however, exist primarily as natural language narratives contained within scientific journal articles. This synthesis information must first be extracted from the text in order to enable analogous synthesis planning methods for inorganic materials. In this work, we present a system for automatically extracting structured representations of synthesis procedures from the texts of materials science journal articles that describe explicit, experimental syntheses of inorganic compounds.  We define the structured representation as a set of linked events made up of extracted scientific entities and evaluate two unsupervised approaches for extracting these structures on expert-annotated articles: a strong heuristic baseline and a generative model of procedural text. We also evaluate a variety of supervised models for extracting scientific entities. Our results provide insight into the nature of the data and directions for further work in this exciting new area of research.
\end{abstract}

\section{Introduction}
\label{sec-intro}
The targeted design and discovery of novel materials remain a key challenge across multiple subfields of chemistry and materials science \cite{curtarolo2013high,jansen2015conceptual,sumpter2015big,gomez2016design}. Accurate, machine-learned predictions of relations between inorganic materials structures and properties have proliferated in tandem with the vast growth of data computed through first-principles methods \cite{meredig2014combinatorial,jain2013commentary,kirklin2015open}, but the progress in predicting and understanding inorganic materials \textit{synthesis} is stagnant by comparison, due to the high cost of producing and tabulating new syntheses.

The syntheses of inorganic materials are available almost exclusively as unformatted natural language text contained within journal articles, and this domain-specific text is often non-trivial to parse \cite{Kim17NatureData}. A broadly-applicable technique for extracting structured representations of inorganic synthesis routes is thus a critical step towards realizing a framework which links synthesis parameters to the properties and structures of produced materials.

In this work, we present a system for automatically extracting structured representations of inorganic synthesis routes. We define these representations of synthesis text based on those used by Kiddon et al. \cite{Kiddon2015MiseEP}. These structures, termed \textit{action graphs}, consist of a set of nodes connected by edges. Nodes represent operations in the synthesis and the arguments associated with each operation. Edges represent the association of an argument with an operation, or denote an argument as having originated from a given operation. Given synthesis procedure text, we first extract individual events in the synthesis using a neural network entity tagger and a set of dependency parse-based heuristics. We then induce edges to compute the sequence of synthesis events. To accomplish this, we employ a simple yet strong baseline method of linking arguments of an event to the previous event, and a modification of the unsupervised generative model for procedural text proposed by Kiddon et al. \cite{Kiddon2015MiseEP}.

Our results indicate that the baseline model which resolves every argument as having arisen in the previous operation out-performs the generative model to induce edges between events in all our evaluation settings. The strong performance of the baseline model hints at the strongly sequential nature of inorganic materials synthesis procedures. Our evaluation also suggests that the bottleneck in extracting high quality action graphs is the event extraction step; our current approach is able to extract about 56\% of the participants of an event (i.e., 56\% of the nodes in the graphs in our test set).

In the following section, we describe related work. This is followed by a description of the action graph extraction task and the graphs themselves. We follow this with a description of our current extraction pipeline. Finally, we present our results and conclusions.

\section{Related work}

\textbf{Materials Science and Chemistry:} The rise of comprehensive materials property and reaction databases has accelerated the development of chemical synthesis planning through the use of first-principles and machine-learning computational techniques \cite{kim2015pubchem,hachmann2011harvard,jain2013commentary,kirklin2015open,lawson2014making,raccuglia2016machine}. Using chemical reaction records retrieved from a database, Grzybowski et al. showed that it is possible to construct a ``universal'' graph of chemistry such that molecules and reactions correspond to vertices and edges, respectively \cite{grzybowski2009wired}: Synthesis action graphs are computed by resolving unitary chemical reactions into action vertices with input and output molecules denoted by directed graph edges. Traversals on this universal chemical reaction graph allow for the optimization of pathways (e.g., for minimizing economic cost) \cite{grzybowski2009wired}, but methods for predicting novel, highly-structured synthesis pathways have remained elusive until very recently. The work by Coley et al. and Segler et al. investigates two complementary problems by learning on historical chemical reaction databases \cite{coley2017prediction,segler2017learning}. Using a neural-network-driven candidate ranking approach, Coley et al. produce a model for predicting organic reaction outcomes \cite{coley2017prediction}. Conversely, Segler et al. approach the opposite problem, using Monte Carlo tree search to predict a synthesis pathway for a given output molecule. Impressively, the results attained by Segler et al. are shown to perform at a level comparable to human-driven organic molecule synthesis planning \cite{segler2017learning}.

While significant strides have been made in computational synthesis planning via the prediction of synthetic action graphs in organic chemistry, these approaches have relied significantly on datasets comprised of historical chemical reactions. Despite efforts to standardize the reporting of chemical and materials science data \cite{murray1999chemical}, \textit{inorganic} materials synthesis routes continue to reside in non-standard form. Several past studies have applied a variety of text extraction techniques to materials science literature using regular expressions, lexicon matches, or human-driven data labelling to extract synthesis information \cite{hawizy2011chemicaltagger,jones2014automatic,young2017data,raccuglia2016machine,ghadbeigi2015performance}, but such approaches are inherently difficult to scale across sub-domains of materials science since new rules and lexicons must be created by domain experts for different types of materials or synthesis techniques. Accordingly, Kim et al. have previously developed automatic methods for extracting aspects of inorganic synthesis routes from natural language text to step in the direction of building comprehensive databases of the syntheses of inorganic materials~\cite{Kim17NatureData,Kim2017ACS}.

\textbf{Natural Language Processing (NLP):} Extracting action graphs relates closely to the problem of extracting event chains in the NLP literature. Formalized by Schank and Abelson \cite{SchankAbelson1977}, this line of work defines domain-specific (typically news wire) structured representations of a prototypical sequence of events. Work by Chambers and Jurafsky \cite{Chambers2009SchemaPart} extracts chain-of-events structures from newswire text using co-occurrence counts of verb-argument pairs and clustering verbs and arguments using mutual information-based similarity metrics. Many extensions to this line of work have been proposed \cite{Balasubramanian2013atscale, Pichotta2014MultiargEvent, Cheung2013ProbFrame}. These approaches are typically either trained with supervised data, or make many domain-specific assumptions which do not carry over to materials science syntheses. Instead, our work is based on that of Kiddon et al. \cite{Kiddon2015MiseEP}, who introduce the notion of action graphs as formalizations of event chains for procedural text (cooking recipes), proposing a generative model to extract these structures. This work has also recently been used to extract action graphs from instructional videos and transcripts \citep{Huang2017LLN}.

\section{Task definition}
\label{sec-prob_def}
\begin{figure}[t]%
\vspace{-0.9cm}
    \centering
    \subfloat[Example of typical synthesis procedure text.]{\fbox{\includegraphics[clip, trim=0.3cm 0.1cm 0.3cm 0cm,width=0.442\textwidth,valign=c]{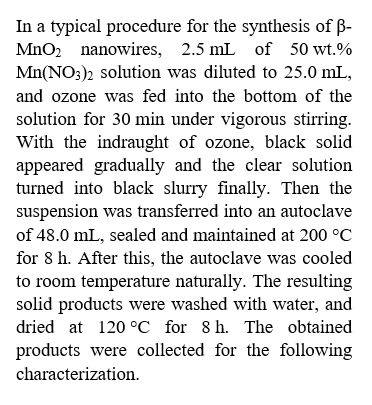}}}%
    ~%
    \subfloat[Possible partial action graph for text.]{\fbox{\includegraphics[clip, trim=8.5cm 9cm 9.5cm 0cm, width=0.4\textwidth, valign=c]{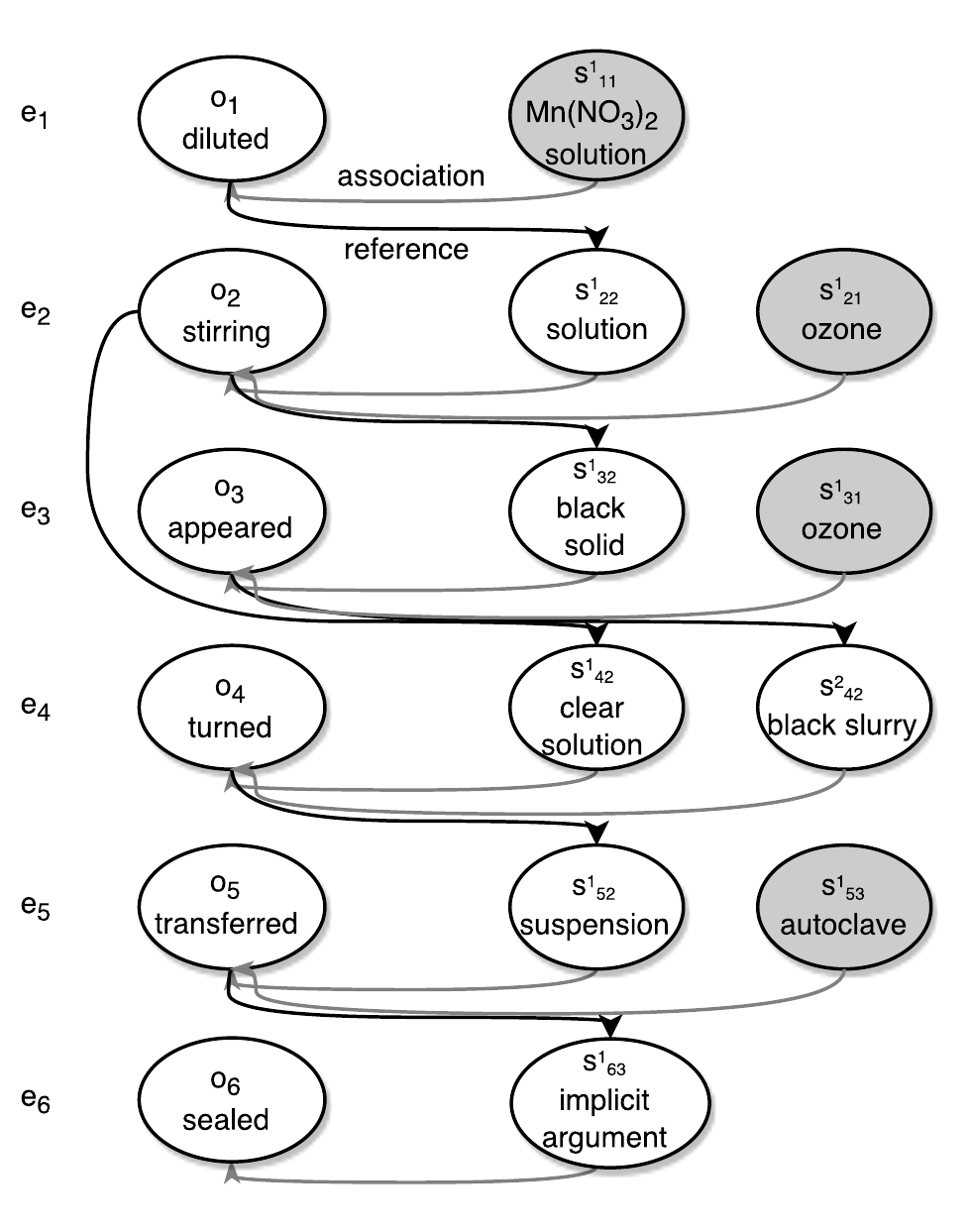}}}
    \caption{Example of a synthesis procedure and the shortened action graph for the synthesis procedure, adapted from Dong et al. \cite{dong2009beta}. The nodes of the graphs are the operations and arguments and the edges represent association between event arguments or reference across events. Ellipsis/missing arguments are dealt by adding ``implicit argument'' nodes. Nodes in gray are lack reference edges and represent ``raw'' nodes.}%
    \label{fig-running_ex}%
\end{figure}
	We aim to extract structured representations of synthesis procedure text, as reported in journal articles, describing inorganic (e.g., hydrothermal, sol-gel, solid state) syntheses of materials. An example synthesis procedure \cite{dong2009beta}, depicted in Figure \ref{fig-running_ex}, may be viewed as a set of events.  Each event consists of an operation and a set of arguments: the arguments may be entities such as the conditions for the operation, apparatuses used, and materials involved. The structure we extract from the text is an \textit{action graph} where the nodes are the operations and their sets of arguments. The edges within an event represent association of an operation with its arguments, and edges between events represent the ``flow'' of arguments as having originated from certain operations.
    
    Extraction of action graph structures from synthesis text presents a two-fold challenge. First, given sentences from scientific articles, it is difficult to extract the correct set of events and arguments. This is compounded by the fact that most sentences also tend to describe multiple events. Indeed, our data indicate that each sentence contains on average two operations. Second, resolving references of arguments between events is non-trivial. As an example, it is necessary to correctly determine the origin of an intermediate material, referred to as a ``black solid,'' when constructing the action graph for the synthesis procedure shown in Figure \ref{fig-running_ex}. In the present setting, the resolution of references requires some domain knowledge since arguments change physically and chemically between events; in this case, the ``black solid'' turned into the ``black slurry''. Another compounding factor is the presence of ellipses or missing arguments; for example, ``sealed'' and ``maintained'' both lack an explicitly mentioned argument, although it is clear from the context that the argument is an ``autoclave.''
    
    \subsection{Action Graph Formalism}
    We define action graphs by modifying the definitions in Kiddon et al. \cite{Kiddon2015MiseEP}. The set of events $E_\mathcal{S}$ in the synthesis procedure $\mathcal{S}$ are represented by $E_\mathcal{S}$. The event set consists of events $e_i$, $E_S= \{e_1=(o_1,\mathbf{a_1})\dots e_n=(o_n,\mathbf{a_n})\}$. Each event is a tuple of the form $(o_i,\mathbf{a_i})$, an operation $o_i$ and a set of typed arguments $\mathbf{a_{ij}}$. Every typed argument is a tuple, $(t_{ij}^{sem}, S_{ij})$. Here $t_{ij}^{sem}$ represents a semantic type of the argument and $t_{ij}^{sem} \in \{{raw-material},{intermediate},{apparatus}\}$ and $S_{ij} = \{s_{ij}^1..s_{ij}^k..s_{ij}^{|S_{ij}|}\}$ is the set of string spans from the text which are instances of the arguments. The string spans $s_{ij}^k$ and the operations $o_i$ therefore represent the nodes of the graph. There are two types of edges, ``association'' edges within events and ``reference'' edges between events (labeled in Figure \ref{fig-running_ex}). Edges of association are represented by the 4-tuple $(o,i,j,k)$ where $o$ identifies the operation and $(i,j,k)$ identifies the string span. Similarly, references edges are made up of a 5-tuple $(o,i,j,k, t^{sem})$. Ellipsis are handled by introduction of ``implicit argument'' nodes. Nodes for raw-materials or explicit apparatus (shown in gray in Figure \ref{fig-running_ex}) lack reference edges.
    
\section{Extraction pipeline}
\label{sec-pipeline}
\textbf{Synthesis text extraction:} Our pipeline begins by extracting paragraph demarcated raw text from PDF documents, using a method described in more detail by Kim et al. \cite{Kim17NatureData,Kim2017ACS}. The plain text articles are extracted using the WatrWorks\footnote{WatrWorks, Text Extraction and Annotation Suite: \url{http://iesl.github.io/watr-works/}} text extraction tool. From this text, we extract the set of paragraphs which are likely to contain synthesis procedures. This is accomplished using a logistic regression classifier trained on word-embeddings and a set of manually designed features \cite{Kim17NatureData}. The word embeddings\footnote{Embeddings: \url{https://github.com/olivettigroup/materials-word-embeddings}} were obtained by training the Word2Vec algorithm on a corpus of materials science research papers \cite{Kim17NatureData}. Multiple paragraphs in a given paper labeled by the classifier as having a synthesis procedure are treated as part of the same synthesis procedure. Sentence and token boundaries are determined with ChemDataExtractor \cite{Swain2016CDE}, which is specifically tailored to processing chemistry-related text.
    
\textbf{Entity extraction:}
Given this synthesis text, our next task is to extract entity mentions from the text. We use the term ``entity mentions'' to denote spans of text that will participate in the experiment, such as \emph{black slurry} or \emph{stirred}. We cast this as a supervised task akin to the classic NLP problem of Named Entity Recognition or Entity Extraction, generating training data by manually annotating a small set of papers for this purpose (see Section \ref{entity-data} for details).

We experiment with the following probabilistic models for entity extraction. Let $x = [x_1, \ldots, x_T]$ be a sentence of input text and $y = [y_1, \ldots, y_T]$ be per-token output tags. Let $D$ be the number of possible labels for each $y_i$. We predict the most likely $y$, given a conditional model $P(y \mid x)$. 

We experiment with two factorizations of $P(y \mid x)$. First:
\begin{equation}
P(y | x) = \prod_{t = 1}^T P(y_t | F(x)),
\label{eq:cond-ind}
\end{equation}
where tags are conditionally independent given some features for $x$. These features could be a binary vector representing each token's membership in e.g. a lexicon, or they could be a dense vector encoded by a deep neural network which takes distributed representations of words as input. In Section \ref{entity-results} we present experiments on the latter, where the deep neural network is either a bidirectional LSTM \citep{lample2016neural}, or a dilated CNN \citep{Strubell17ConvNER}.

We also consider a linear-chain CRF model that couples all of $y$ together, enforcing constrants between different labels during prediction:
\begin{equation}
P(y | x) = \frac{1}{Z_x} \prod_{t = 1}^T \psi_t(y_t | F(x)) \psi_p(y_t,y_{t-1}),\label{eq:crf}
\end{equation}
where $\psi_{t}$ is a local factor, $\psi_p$ is a pairwise factor that scores adjacent tags, and $Z_x$ is the partition function~\citep{lafferty2001conditional}. Prediction in this model is performed using the Viterbi algorithm. In Section \ref{entity-results} we experiment with models where $F(x)$ is encoded by a bidirectional long short-term memory (LSTM), a pre-trained word embedding, a binary vector constructed from hand-engineered features based on linguistic analysis such as syntactic parses of the sentence and part-of-speech tags, and combinations thereof.

\begin{figure}[t]%
    \vspace{-0.5cm}
    \centering
    \subfloat[Example result of running Stanford CoreNLP dependency parser.]{\includegraphics[clip, trim=6.5cm 12cm 4cm 1cm,width=0.7\textwidth,valign=b]{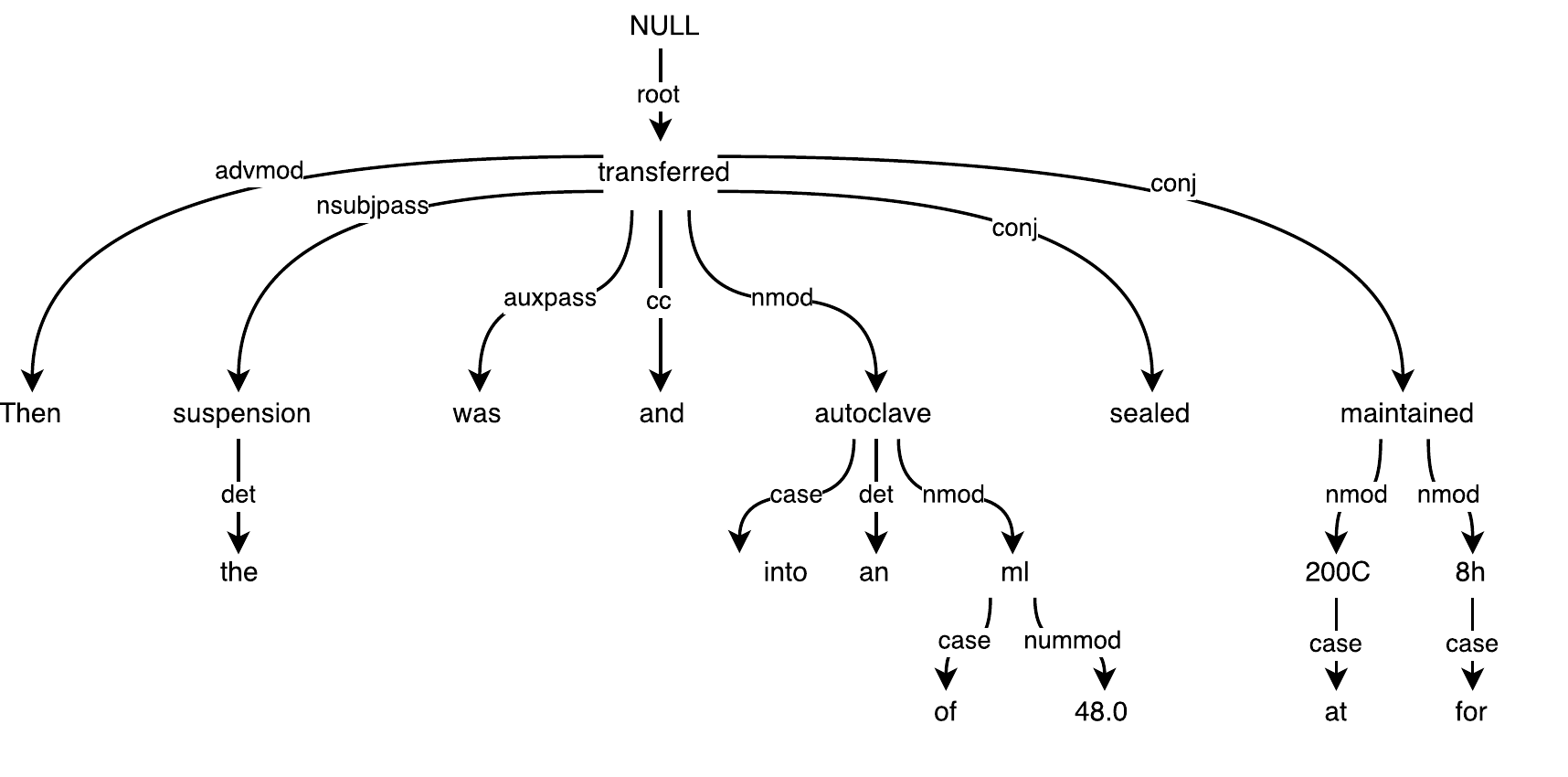}}%
    ~%
    \subfloat[Tokens grouped by relations to the root token]{\includegraphics[clip, trim=9cm 15cm 9cm 1cm, width=0.3\textwidth, valign=b]{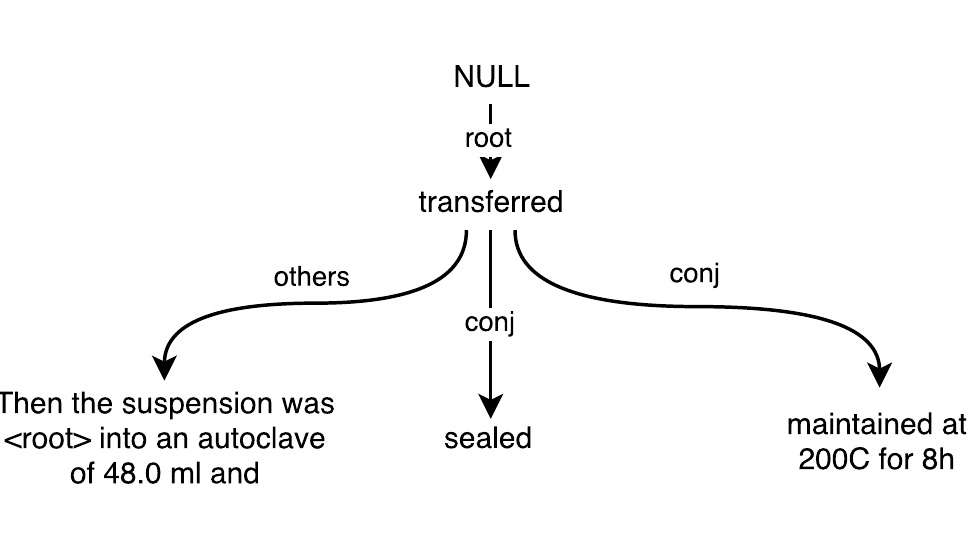}}
    \caption{An example illustrating use of dependency graphs to split a sentence consisting of multiple events into a set of separate phrases for each event. Phrases with the \texttt{conj} relation to the root word are broken off as potentially separate events.}%
    \label{fig-deps_simplification}%
\end{figure}

    
\textbf{Event extraction:} Once we have extracted entities, we must combine them into events. Since sentences from the synthesis text often describe multiple events we break the sentence into separate event phrases by applying heuristic rules on the dependency parse of the sentence. We obtain dependency parses from the Stanford CoreNLP dependency parser \citep{manning-EtAl:2014:P14-5}. Given the parse, the most important of these heuristics breaks every phrase, whose head token links to the sentence root with a \texttt{conj} relation. All other tokens are associated with the root and constitute the main phrase. Each split phrase and the main phrase is considered to be an event.  In the case of no \texttt{conj} relation to the root, the whole sentence is considered a single event. We apply these heuristics only to sentences with multiple operations. On identifying sentence segments representative of events all the tokens tagged as operation, raw-material, intermediate or apparatus are extracted and treated as nodes of the particular event. Figure \ref{fig-deps_simplification} denotes the illustration of an example. Implicit argument nodes are added to events lacking an argument of the apparatus or intermediate type.

	\textbf{Edge induction:} Following the extraction of individual events, we then induce the reference edges from argument nodes of intermediate materials to operations. These edges denote the operation that gave rise to a given intermediate material. In inducing these edges we attempted two approaches. The first of these approaches assumes that every intermediate material is derived from the previous operation. This approach forms our baseline (Sequential model) and as we elaborate in Section \ref{sec1-results} turns out to be a very strong baseline. For example, in Figure \ref{fig-running_ex}, the sequential model would link ``black solid'' to ``stirring'' and link ``black slurry'' to ``appeared''. The baseline therefore assumes a strong sequential structure for the events and assumes text order to be the correct order in resolving entity origins. The other approach we tried an unsupervised generative model proposed by Kiddon et al. \cite{Kiddon2015MiseEP}. This model, trained with a hard-EM algorithm explicitly models the structure of procedural text and attempts to connect arguments to one of previous event operations. However, as we present in Section \ref{sec1-results} the generative model did not perform as well on out dataset, we think, mainly due to the strongly sequential nature of the data itself. Next, we describe our approach to evaluating the action graphs and the data that we use in analysis of our results.

\section{Experimental Results}

We present experimental results on two components of our action graph extraction pipeline: entity extraction, in which we identify mentions of scientific entities in the text, and action graph extraction, the final output of our pipeline. 

\subsection{Entity Extraction}

We evaluate six models for entity extraction: Three baseline linear-chain CRF models using different sets of linguistically motivated features, three models which perform logistic regression with token representations encoded by different neural network architectures, and a combined structured neural network model which takes the token representations encoded by a neural network as the logits for a linear-chain CRF (Eqn. \ref{eq:crf}). 

Specifically, the {\bf CRF-ling} model is a CRF  where the feature representation for each token is a binary feature vector encoding various linguistic features of the token and lexicon membership. Linguistic features include: part-of-speech tag, lemma, stem, syntactic dependency label, 
and whether the token contains numbers and its capitalization. The lexicons (dictionaries) include: stop words (non-content words such as \emph{the}, or \emph{that}), entries in ChemDB \citep{chen2005chemdb}, and hand-built lists of typical operations, conditions, entities, amounts, descriptors and apparatuses. Annotations were extracted using Stanford CoreNLP and NLTK. The {\bf CRF-hand} model is a CRF which takes as per-token features the concatenation of its pre-trained word embedding with a binary feature vector indicating whether the token matched any of a set of hand-crafted rules over parse trees. {\bf CRF-both} uses as features the concatenation of all the features from the CRF-ling and CRF-hand models. 

We also evaluate three neural network-based models: a three-layer dilated CNN ({\bf DCNN}), a bidirectional LSTM ({\bf Bi-LSTM}), and a bidirectional LSTM used in conjunction with inference in a CRF ({\bf Bi-LSTM-CRF}). All three models take the same input features for each token: its pre-trained word embedding and a learned embedding representing the word shape (capitalized, all caps, all lowercase, etc.) Rather than performing structured prediction in a CRF, the CNN and Bi-LSTM models all predict tokens using a logistic regression classifier (Eqn. \ref{eq:cond-ind}) whose features are the token representations encoded by the neural network. Each neural network is trained end-to-end with the classifier. The Bi-LSTM-CRF performs structured inference in a CRF with inputs encoded by a Bi-LSTM. This model is also trained end-to-end with the partition function and its gradient computed using the forward-backward algorithm.

\subsubsection{Data \label{entity-data}}
We manually annotated text extracted using the PDF processing pipeline described in Section \ref{sec-pipeline} from the methods section of 42 materials science papers from a variety of experimental materials science journals. We define a set of 18 entity type labels with types reflecting typical roles present in a synthesis route, such as materials, operations, targets and amounts: {\sc target}, {\sc material}, {\sc descriptor}, {\sc amt\_unit}, {\sc cnd\_misc}, {\sc cnd\_unit}, {\sc intermed}, {\sc operation}, {\sc number}, {\sc amt\_misc}, {\sc prop\_unit}, {\sc prop\_type}, {\sc prop\_misc}, {\sc synth\_aprt}, {\sc char\_aprt}, {\sc brand}, {\sc meta}, and {\sc ref}. See \cite{Kim17NatureData} for more details. In all, including tokens labeled as non-entities, the dataset at the time of writing consists of just under 10,000 tokens. We divided the 42 papers into 29 training (70\%) and 13 test documents. In all models that use optimization or other hyperparameters, we simply use default settings (CRF models) or settings derived from tuning on data from a different domain (neural models).

We use the BILOU (Begin, Inside, Last, Outside, Unit) segment boundary encoding, augmenting each token's label to indicate its position in the span of tokens making up the entity (for example, the two tokens in the span \emph{black solid} might have the labels {\sc B-intermed}, {\sc I-intermed}). Previous work has found this encoding to result in improved performance \citep{ratinov2009design}. 

\subsubsection{Evaluation \label{sec-eval}}

We evaluate our entity extraction models using segment-wise precision, recall and F1, the harmonic mean of precision and recall. Since a single entity mention may consist of multiple tokens (for example, the entity \emph{black solid} consists of the tokens \emph{black} and \emph{solid}), we mark a prediction as correct only if every token in the entity is correctly labeled. In other words, there is no partial credit. We compute true positives as correctly predicted entities; False positives are entities whose first token predicted by our system does not align with the first token of a labeled entity; and similarly, false negatives are labeled entities whose first token our system failed to predict. 

\subsubsection{Results \label{entity-results}}

Table \ref{entity-table} presents our entity extraction results. Table \ref{entity-p-r-f} lists precision, recall and F1 achieved by each of the models. Among the CRF models using linguistic and hand-engineered features, we find that the CRF-hand model, which uses hand-engineered features over parse trees combined with word embeddings, out-performs the model which is given a wide array of linguistic features, including parse information. CRF-both, which combines both sets of features, outperforms each individual model, achieving an overall F1 of $73.52$. Whereas the individual models are more precision-biased, CRF-both achieves much higher recall than either of the individual models though its precision suffers slightly, leading to the best overall F1 among these models.

All of the neural network-based models outperform the best feature-based CRF model, even those that predict using simple logistic regression. Among the models using logistic regression, the DCNN performs the best with an F1 of $77.50$. The best performing model of all, by an insignificant margin, is the Bi-LSTM-CRF with an F1 score of $77.57$, though its performance is comparable with the DCNN. This result is on par with those reported in \citet{Strubell17ConvNER}, which show that the DCNN is a higher quality neural model for encoding rich token representations which incorporate wide context, and thus which do not require the structure from a CRF in order to form accurate predictions. 

\begin{table}[t]
\centering
\scalebox{0.85}{
\subfloat[Precision, recall and F1 scores of the entity extraction models. Neural network models out-performed feature engineered CRFs across the board. \label{entity-p-r-f}]{
\begin{tabular}{|lrrr|} \hline
Model & Precision & Recall  & F1\\ \hline \hline
CRF-ling    & 76.98   & 67.41 & 71.88\\
CRF-hand & 75.59   & 69.32 & 72.48\\
CRF-both  & 74.97   & 72.12 & 73.52\\\hline
DCNN & \textbf{77.85}   & 77.16 & 77.50\\ 
Bi-LSTM     & 74.25   & 77.83 & 76.00\\
Bi-LSTM-CRF   & 74.64   & \textbf{80.74}  & \textbf{77.57} \\\hline
\end{tabular}
}
\quad
\subfloat[Breakdown of F1 score by label in the CRF, ID-CNN and Bi-LSTM-CRF models (abbreviated Bi-L-CRF) for labels with more than 10 annotated entities.\label{entity-class-breakdown}]{
\begin{tabular}{|lrrr|r|} \hline
Label & CRF-both & Bi-L-CRF & DCNN & Support\\\hline \hline
target    & 27.91 & 32.65   & 36.36    & 36\\
material  & 74.19  & 80.16   & 82.11    & 95\\
descriptor  & 57.58 & 62.03   & 62.64    & 64\\
amt\_unit & 83.48 & 83.48   & 83.48     & 45\\
cnd\_misc & 69.23  & 74.63  & 72.73     & 42\\
cnd\_unit & 95.10 & 94.52   & 93.06     & 91\\
intrmed   & 64.58 & 73.91   & 75.12     & 29\\
operation & 80.95  & 82.76    & 82.55     & 146\\
number    & 87.71 & 91.89    & 88.67     & 114\\ \hline
\end{tabular}
}
}
\caption{Evaluation of the entity extraction models \label{entity-table} } 
\end{table}

Table \ref{entity-class-breakdown} lists F1 scores for the CRF-both, Bi-LSTM-CRF and DCNN models broken down by label, for labels which occurred more than ten times in the test data, along with the occurrence count for each label. There is a clear trend, where all the models tend to perform better on labels which have more support (we expect the distribution in the test data mimics that of the training data). For labels with fewer examples such as {\sc target} and {\sc intermed}, the DCNN tends to perform better. 




\subsection{Action Graph Extraction}
\subsubsection{Data}
\label{sec1-data_description}
Our current dataset for this task consists of 240 materials science journal articles which were acquired using publisher approved APIs and text-mining interfaces. Details of this process are described in Kim et al. \cite{Kim17NatureData}. Fifteen of these articles were annotated with the action graph structures. The 15 articles were selected to evenly represent different sub-types of inorganic syntheses (e.g., hydrothermal, sol-gel, solid state), and to ensure that annotated articles contained explicit synthesis descriptions of inorganic materials. Annotation was performed using an in-house web application.

The extraction pipeline was run on the entire set of 240 documents. Edges were induced by using the generative model described Kiddon et al. \cite{Kiddon2015MiseEP} and by the sequential baseline model. We evaluate the results of both approaches by evaluating the predicted graphs for the 15 test cases. Next we describe our evaluation strategy for the action graphs.

\subsubsection{Evaluation}
\label{sec1-eval_strategy}
Evaluating the extracted graphs involves comparing predicted and ground truth graphs. Our evaluation strategy involved a two step process. Given the two graphs we first align the sets of nodes in both graphs. Once aligned, we score individual edges by checking for its presence in the annotation, and compute an F1 score. 

The alignments of nodes are made by means of exact token index matches between the nodes in the predicted graph and the annotated graphs. Using these alignments we report the fraction of nodes in the predicted graph which were aligned and unaligned as alignment scores. Since we match nodes based on token indices in the source text, we currently do not align implicit argument nodes, marking them as unaligned. 

Given aligned nodes, we find true positive, false positive and false negative edges. We define these metrics as follows: True positives are edges present in the predicted graph and the annotated graph; false positives are edges present in the predicted graph but not in the annotated graph; and false negatives are edges not present in the predicted graph but present in the annotated graph. Based on these metrics we compute micro-averaged precision, recall and F1 on our test set.


\subsubsection{Results}
\label{sec1-results}
\label{sec-exp and results}
We evaluate the induced edges when the nodes have been generated from running our entire pipeline end to end (End-to-end evaluation in Table \ref{tab-eval}) and also in the case where the nodes have been induced from the annotations (Perfect node segmentation in Table \ref{tab-eval}). The second case simulates the case of having a perfect set of operation and argument nodes. We do this so as to analyze errors made by different stages of the pipeline (i.e., the event extraction and the edge induction models). 

For both of the above cases, we also perform evaluations under two settings. In the first setting we ignore edges between unaligned nodes; We call this \emph{Setting 1} in Table \ref{tab-eval}. In the second setting we systematically penalize all edges which have one or both nodes unaligned in the predicted graph as being False Positives; We call this \emph{Setting 2} in Table \ref{tab-eval}. We do not distinguish between reference and association edges. 

\begin{table}[t]
\centering
\scalebox{0.95}{
\begin{tabular}{|ccccccccc|}
\hline
\bf Model             & \bf Aligned    & \bf Unaligned    & \bf Precision  & \bf Recall  & \bf F1     & \bf Precision  & \bf Recall  & \bf F1     \\ \hline
\multicolumn{3}{|c}{End-to-end evaluation}     & \multicolumn{3}{c}{Setting 1} & \multicolumn{3}{c|}{Setting 2} \\ \hline
Sequential        & 39.85\%    & 30.95\%      & \textbf{73.04}      & \textbf{94.38}   & \textbf{82.35}  & \textbf{27.10}      & \textbf{27.91}   & \textbf{27.50}  \\ 
Probabilistic     & 39.85\%    & 30.95\%      & 68.38      & 89.89   & 77.67  & 25.81      & 26.58   & 26.19  \\ \hline
\multicolumn{3}{|c}{Perfect node segmentation} & \multicolumn{3}{c}{Setting 1} & \multicolumn{3}{c|}{Setting 2} \\ \hline
Sequential        & 63.80\%    & 0\%          & \textbf{99.29}      & \textbf{99.29}   & \textbf{99.29}  & \textbf{99.29}      & \textbf{92.36}   & \textbf{95.70}  \\
Probabilistic     & 63.80\%    & 0\%          & 95.36      & 95.36   & 95.36  & 95.36      & 88.70   & 91.91  \\ \hline
\end{tabular}
}
\caption{Evaluation of action graph extraction in terms of precision, recall and F1. End-to-end evaluation uses entities as predicted by our system, while perfect node segmentation uses the annotated entities. \emph{Setting 1} ignores all edges where at least one node is unaligned. \emph{Setting 2} penalizes edges where at least one node is unaligned as false positives. \label{tab-eval}}
\end{table}

The strong performance of the sequential model in both evaluation settings, in the end-to-end and the perfect node segmentation cases indicate the strong sequential nature in the data. Almost all intermediates derive directly from the previous operation. This is particularly apparent in the perfect node segmentation cases where the baseline has a F1 scores greater than 0.95. This seems to shed light on the nature of the data itself. 

The alignment scores in the end-to-end case in both evaluation settings hints at the problems in the extraction of the action graph structures. The results above indicate that we currently extract about 56\%\footnote{Explicit nodes aligned: $0.3985/(0.3985+0.3095)=0.5628$} of all non-implicit argument nodes in the predicted graphs. The major challenge in this extraction task seems to be being able to identify individual operations and their arguments. Next we present some conclusions and approaches we plan to pursue in future work.

\section{Conclusions}
\label{sec-conclusion}
In this work, we present models for extracting action graph structures from materials science synthesis procedures without access to any labeled target structures. Our experimental results highlight: (1) neural network models with word embedding features significantly outperform classic linear chain CRF model with manually designed features on the NER task on material science text despite a small training dataset. (2) merely resolving every intermediate as having arisen from the previous operation leads to very strong scores on our current dataset and evaluation metrics in the action graph extraction task, and (3) the major hurdle to extracting action graphs from the synthesis text is the accurate identification of its operations and arguments. Future work will explore improving entity and event extraction despite data scarcity, modeling event extraction as e.g. unsupervised n-ary relation extraction \citep{Verga2016CompUschema}, and end-to-end training of a single neural network which jointly models entities, events and actions.

\section*{Acknowledgements}
We thank Kathryn Ricci and Zach Jensen for help with annotation and pipeline engineering. The authors would also like to acknowledge funding from the National Science Foundation Awards 1534340 (DMREF) and IIS-1514053, support from the Office of Naval Research (ONR) under Contract N00014-16-1-2432, the MIT Energy Initiative, the UMass Center for Data Science and the Center for Intelligent Information Retrieval, and in part by the Chan Zuckerberg Initiative under the Scientific Knowledge Base Construction project. E.K. was partially supported by NSERC, and E.S. was supported in part by an IBM Ph.D. Fellowship Award. The authors would also like to acknowledge support from seven major publishers who provided the substantial content required for our analysis.

\bibliographystyle{unsrtnat}
\bibliography{material_sci_age}
\end{document}